# Functional mimicry of Ruffini receptors with Fiber Bragg Gratings and Deep Neural Networks enables a bio-inspired large-area tactile sensitive skin

Luca Massari [1,2*], Giulia Fransvea [1,2,3*], Jessica D'Abbraccio [1,2], Mariangela Filosa [1,2], Giuseppe Terruso [1,2], Andrea Aliperta [1,2,3], Giacomo D'Alesio [1,2], Martina Zaltieri [1,2,4], Emiliano Schena [4], Eduardo Palermo [5], Edoardo Sinibaldi [6**], Calogero Maria Oddo [1,2**]

[1]The BioRobotics Institute, Polo Sant'Anna Valdera, Scuola Superiore Sant'Anna, Viale Rinaldo Piaggio 34, 56025 Pontedera, Italy.

[2]Department of Excellence in Robotics & AI, Scuola Superiore Sant'Anna, 56127 Pisa, Italy

[3]ARTES4.0 Competence Center on Advanced Robotics and enabling digital TEchnologies and Systems, 56025 Pontedera, Italy

[4]Research Unit of Measurements and Biomedical Instrumentation, Department of Engineering, Università Campus Bio-Medico di Roma, Via Alvaro del Portillo 21, 00128 Roma, Italy.

[5]Department of Mechanical and Aerospace Engineering, Sapienza University of Rome, Via Eudossiana 18, 00184 Roma, Italy (EP).

[6]Istituto Italiano di Tecnologia, Via Morego 30, 16163 Genoa, Italy (EDS).

* These authors contributed equally

** Corresponding authors: calogero.oddo@santannapisa.it, edoardo.sinibaldi@iit.it

# Abstract

Collaborative robots are expected to physically interact with humans in daily living and workplace, including industrial and healthcare settings. A related key enabling technology is tactile sensing, which currently requires addressing the outstanding scientific challenge to simultaneously detect contact location and intensity by means of soft conformable artificial skins adapting over large areas to the complex curved geometries of robot embodiments. In this work, the development of a large-area sensitive soft skin with a curved geometry is presented, allowing for robot total-body coverage through modular patches. The biomimetic skin consists of a soft polymeric matrix, resembling a human forearm, embedded with photonic Fiber Bragg Grating (FBG) transducers, which partially mimics Ruffini mechanoreceptor functionality with diffuse, overlapping receptive fields. A Convolutional Neural Network deep learning algorithm and a multigrid Neuron Integration Process were implemented to decode the FBG sensor outputs for inferring contact force magnitude and localization through the skin surface. Results achieved 35 mN (IQR = 56 mN) and 3.2 mm (IQR = 2.3 mm) median errors, for force and localization predictions, respectively. Demonstrations with an anthropomorphic arm pave the way towards AI-based integrated skins enabling safe human-robot cooperation via machine intelligence.

# Introduction

Collaborative robots, or Cobots, should have the capability to interact with humans in a shared workspace [1] in different scenarios, ranging from industrial production, transportation and delivery of goods, up to medical aid and rehabilitation [2–4]. Workers and Cobots are expected to physically cooperate in unstructured common spaces and with the ongoing Industry 4.0 transition, the machine will no longer be considered a potential substitute but rather a companion, assisting and complementing human abilities in performing a wide range of tasks.

Current Cobots typically integrate sensing technologies for contact detection, such as force/torque sensors [5], complementing proximity identification [6,7]. These technologies lead to constraints on modularity, scalability and retrofit to the installed non-collaborative machines and require low inertial and payload configurations too. Since their presence might be harmful for humans, robots still operate inside closed cages, and they are kept separated from workers in most processes. In this domain, accidental or voluntary contact might occur; thus, the availability of intelligent sensing systems would be essential toward a coexistence in unstructured environments. A robot able to sense, categorize and respond to touch throughout its body, ideally mimicking the human sensory performance, might lead to more meaningful and intuitive interactions [8], enhanced flexibility, reproducibility, productivity, and risk reduction.

Thus, safe physical co-operations and interactions with the surroundings depend on the availability of tactile feedback, being touch the sensory modality that enables humans to gather a variety of haptic information about the external world by exploring object properties through contact and manipulation [9,10]. The main families of human mechanoreceptors that provide the brain with short-latency feedback to enable closed-loop sensorimotor control are innervated by myelinated fibers and are classified in categories depending on end-organ morphology and positioning with respect to the skin structure, determining the neural encoding of the mechanical input signal [11]. Mechanoreceptors are defined as Slowly Adapting (SA) or Fast Adapting (FA) depending on their temporal response, with SA units responding to sustained indentations and FA ones mainly encoding stimulation transients. Type 1 (surface-located) or type 2 (deeply-located) classes instead reflect the positioning referred to the epidermal layer, with impact mainly on the receptive field and definition of the borders. Specifically, type 1 units have smaller and well-defined fields with multiple hot spots (i.e., the regions with maximal sensitivity within the receptive field), whereas type 2 ones present blurred sensitive regions with single larger spots [12,13]. The combination of temporal response and spatial properties returns four main classes, namely FA1 (Meissner corpuscles), SA1 (Merkel cells), FA2 (Pacinian corpuscles) and SA2 (Ruffini corpuscles) [14]. The integration of multiple mechanoreceptor spiking outputs gives rise to perceptual functions in the brain, such as the ability to determine the location and the magnitude of physical contact through the skin [15]. These physiological properties, together with explanatory mechanical models describing interactions with soft materials for tactile sensing [16,17], became a source of bioinspiration for the present study.

An ideal bio-inspired artificial skin should consist of tactile sensors distributed over large curved areas, able to solve tactile stimuli with millimetric localization, milliNewton magnitude sensing, and millisecond temporal accuracy [10,18,12]. Artificial skins should also be soft [19], stretchable [20–23], lightweight [24], conformable [25] and with minimal wiring encumbrance. Therefore, the application of soft components for tactile feedback is pivotal for both their high flexibility and their intrinsic compliance to enable safe interactions.

### Soft e-skin for sensorizing large areas of robot bodies

In the last decade, remarkable sensing patches leveraging on different transduction mechanisms have been presented [26–31] and soft e-skin solutions have been developed for application in collaborative anthropomorphic robots [32] and neuro-prosthetics [33–37]. However, artificial skins are not yet an integral component of robotic technologies unlike vision sensors [38]. Conventional tactile and proximity sensors have been usually developed using bulky and rigid components. Limited flexibility, deformability and adaptability to unconstrained environments prompted, so far, the usage of robots in confined spaces. On the other hand, the emerging class of soft sensing devices, provided with deformable substrates such as polymers, gels and fluids [39–41] in combinations with miniaturized sensors [42], suggests an approach for the growing call for flexibility.

Covering the whole body of anthropomorphic robots with soft and curved sensing components is a major challenge that is being addressed by state-of-the-art technologies. As an example, the HRP-2 humanoid embeds a multi-modal artificial sensor system which mimics the functional layers of the human skin by means of an optical proximity sensor, a 3-axis accelerometer, a normal force sensor and a temperature sensitive element [43]. Similarly, the iCub integrates a soft skin endowed with 3-axis force sensitive elements [44]. A step towards highly accurate and reliable soft robotics has been presented with a combination of advanced electronic functionalities and skin-like stretchability to develop a 347-element transistor array for force mapping [45]. Another soft e-skin was recently based on a capacitor array within a polyurethane matrix and tested when mounted onto the end-effector of a robotic arm. This system achieved both normal and shear forces estimation in real-time with high sensitivity and excellent cycling stability, with the tracked signal being also used to control and stop the system during predefined tasks [46].

Recently developed pressure sensitive e-skins [47] presented stretchability and conformability, even when simultaneously detecting more than two stimuli [48]. Nevertheless, their row-column *taxels* addressing modality entails major wiring issues, especially when integrating a high number of sensors over complex curved shapes for applications in humanoid robots, human-machine interfaces and healthcare machines.

## Sensor data and AI to enable human-robot cooperation

The ongoing Industry 4.0 manufacturing framework and, in particular, personalized mass production, enabled by Human-Robot Cooperation (HRC), require the enrolment of Cobots, which are able to dynamically change their pre-programmed tasks and share the workspace with human operators [49]. Traditional control methods do not often match the needs of a sensor-based, flexible and integrated solution, therefore HRC applications may benefit from Deep Learning (DL) algorithms to overcome the current limitations in modelling and mimicking human activities [50].

The viability of DL has been proven in several HRC applications, such as computer vision [51], object identification [52], speech and body posture recognition [53]. Furthermore, the potentiality of DL with the implementation of Convolutional Neural Networks (CNN) has also been investigated toward the development of electronic skins with biologically inspired tactile properties, thus mimicking the mechanoreceptors of the human skin [54].

This work presents a bio-inspired sensitive skin, able to detect external tactile stimuli in terms of both contact location and magnitude, thus providing robots with the ability to dynamically interact with the environment. Such an intelligent artificial skin will foster robots awareness and understanding of the surroundings in dynamic tasks, by means of an integrated tactile sensor array inspired to some extent by the large and single-hot-spot receptive fields of Ruffini SA2 corpuscles [11] combined with DL strategies. The degree of bioinspiration is mainly achieved with the deep positioning and shape of FBG sensors in the soft polymeric materials, to emulate the spatial properties of the receptive fields of fusiform SA2 units. The developed curved modular sensorized e-skin patches integrating photonic FBG sensors allow endowing the whole body of collaborative robots with tactile sensing capabilities (Fig. 1a).

An outstanding feature of the FBG technology is the possibility to inscribe multiple sensing gratings within one single optical fiber core, each associated with its nominal wavelength $\lambda_B$ and fully customizable in terms of length (from 1 mm up to 20 mm) and placement (along the fiber). This asset allows minimizing the number of necessary communication channels, and therefore, reduces wiring management issues. The presented e-skin integrated with FBG tactile sensors advances the state-of-the-art by enabling the complete coverage of the Cobot surface with a single wiring element connecting multiple wavelength-separated transducers. Grounding on the Ruffini-like sensor spatial outputs and DL methods, each patch demonstrated to efficiently decode both the magnitude and the localization of the contact force distributed over the large-area artificial skin. This goal was achieved by leveraging on the overlapping receptive fields of the FBGs, which were further interpreted by means of AI strategies.

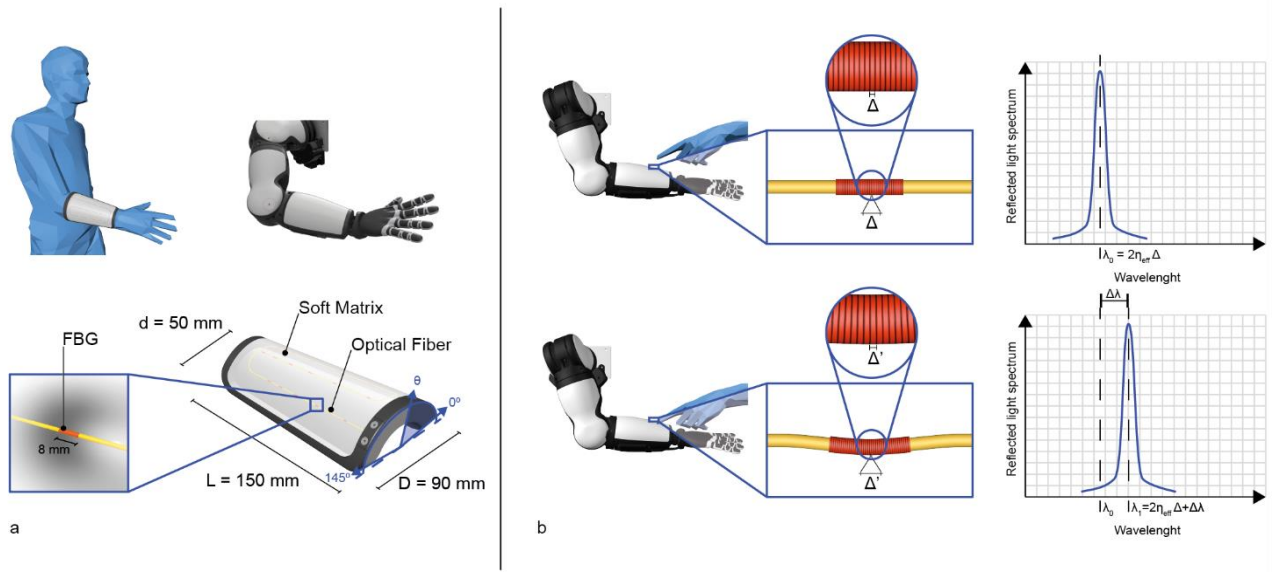

*Fig. 1 | Artificial skin integrating FBG transducers.* a) Sensitive skin patch and its characteristic dimensions; b) FBG Working principle showing how a strain applied to the grating is encoded in peak wavelength shift of the reflected light spectrum.

# Results

### e-skin mimicking spatial properties of Ruffini corpuscles

The presented artificial skin aimed at emulating the human skin functionality by embedding FBGs within a soft polymeric substrate that conveys the applied load to the optical sensors. Specifically, it imitated to some extent the functional role of the human SA2 afferents, through the cross-talk of neighbouring sensors with overlapping receptive fields each having a single large hot-spot to achieve contact localization and force estimation (Fig. 2a-b).

The FBG positioning along the optical fiber implemented a varying spatial density, in order to more coarsely reproduce that of the human mechanoreceptors in the forearm, which increases from the elbow to the wrist [13,55].

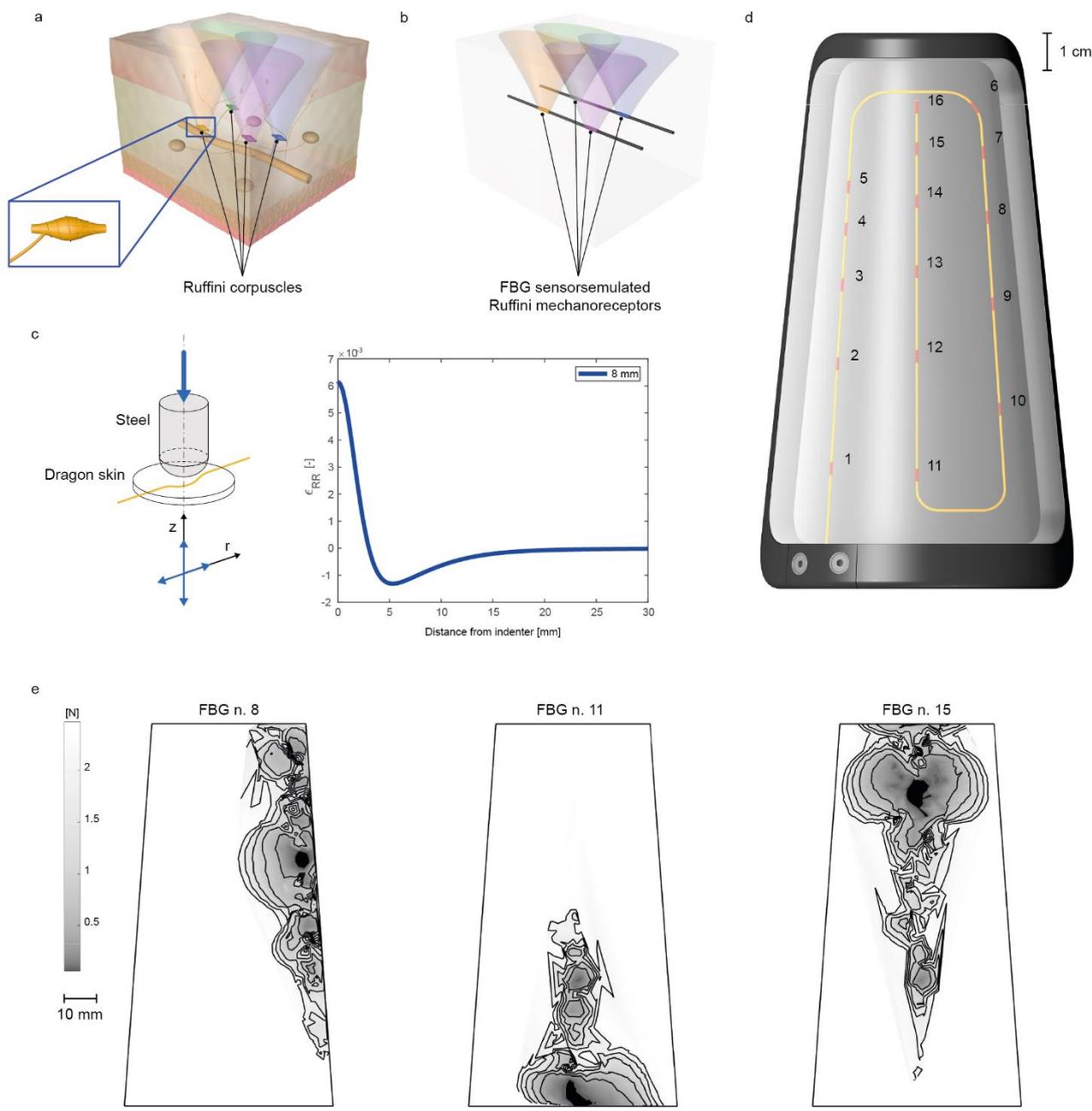

*Fig. 2. Receptive fields and cross-talk of the mechanotransducers embedded in the skin.* a) Human skin with positioning and receptive fields of Ruffini corpuscles highlighted; b) Artificial skin with FBGs; c) Strain (RR component) suffered by the encapsulation material at 50 mN; d) Optical fiber path and positioning of the FBGs in the skin; e) Receptive fields of 3 FBGs characterized by indenting with a hemispherical probe (11 mm diameter) and plotting the force required to achieve a 0.02 nm wavelength variation, corresponding to 10 times the threshold used for contact detection.

Therefore, the developed artificial skin exhibits a bioinspired variable transducer density which was higher closer to the wrist and decreased towards the elbow (with distances between neighbouring FBGs ranging from 12.9 mm to 24.5 mm, Fig. 2d). The FBG technology is suitable for integrating a mesh of transducers since its wavelength-multiplexing functioning guarantees low wiring bulkiness (Fig. 2d). The thickness of the polymeric artificial forearm soft cover, encapsulating the optical fiber in the medial plane, was 8 mm to achieve a trade-off between FBG sensitivity and receptive field size (Fig. 2c) [56] resulting in the triangulation of neighbouring sensors that enabled simultaneous contact localization and force reconstruction via DL methods. This was modelled via FEM analysis with a parametric sweep (in the range 4 mm - 12 mm) on the polymer thickness while applying a load on its top surface (Fig. 2c, boundary conditions and simulation details are given in the Methods section). The simulation results confirmed that the radial RR component of the strain tensor measured in the centre of the polymer layer decreases with the thickness; conversely, the receptive field size increases (Extended Data Fig. 1). The application of loads through calibrated Von Frey hairs to characterize the sensitivity of the skin in human-robot interaction resulted in a sigmoid contact detection rate (Extended Data Fig. 2b; a=2.2 $mN^{-1}$, b=12.4 mN, see code published in the Code Ocean repository for details) for increasing microfilament diameters, with 50.6 mN 75% probability threshold.

Raw FBG sensor data showed the activation of neighbouring FBGs (Fig. 2d), depending on the location and the intensity of the force applied via a hemispherical probe controlled with a mechatronic platform, with 15.9 $mm^2$ median area (9.0 $mm^2$ 1st quartile, 20.9 $mm^2$ 3rd quartile) of the hot spot across all FBGs (reported in black in Fig. 2e). In particular, wavelength changes are correlated with the distance/magnitude from/of the applied load (Fig. 3).

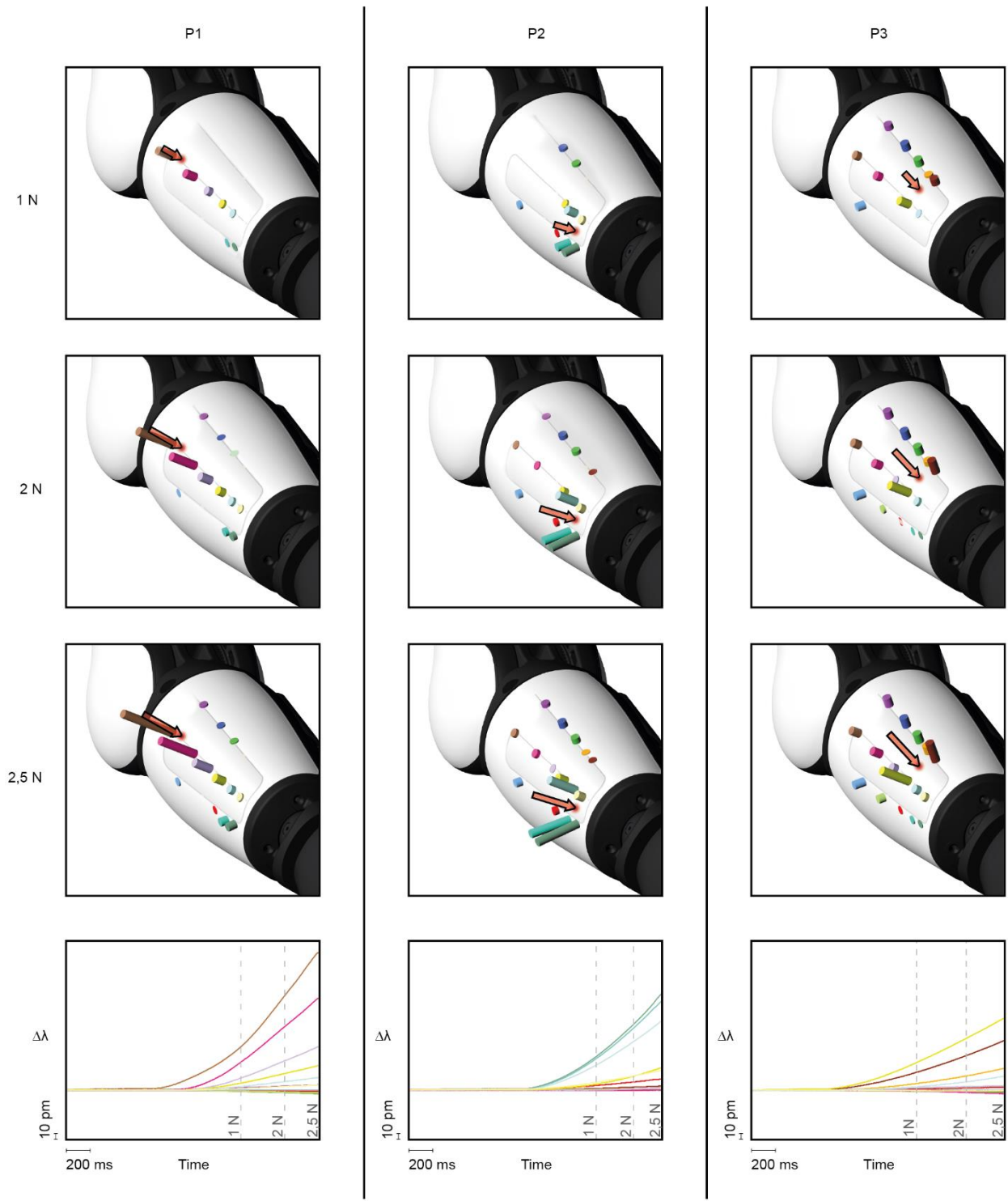

*Fig. 3 | Cross-talk among neighbouring FBGs.* The columns show three different locations of the load, while the rows show three different forces (1 N, 2 N, 2.5 N). The bottom graph shows the raw wavelength variation for all the 16 FBGs as a function of time.

## Deep Learning for contact force and localization inference

Considering the fine manipulation activities that are typically categorized as gentle touch [57–59], both CNN and Multilayer Perceptrons (MLPs) were trained (Fig. 4) based on a series of force-controlled indentations, up to 2.5 N, performed on the surface of the skin in random positions uniformly distributed over 120 mm in length (Y) and 90° rotation span (R) around the elbow-wrist Y-axis (Fig. 5b). The CNN reconstructed the applied external force and a system of 4 MLPs localized the contact source, both relying on the 16 FBG readouts which were the network inputs.

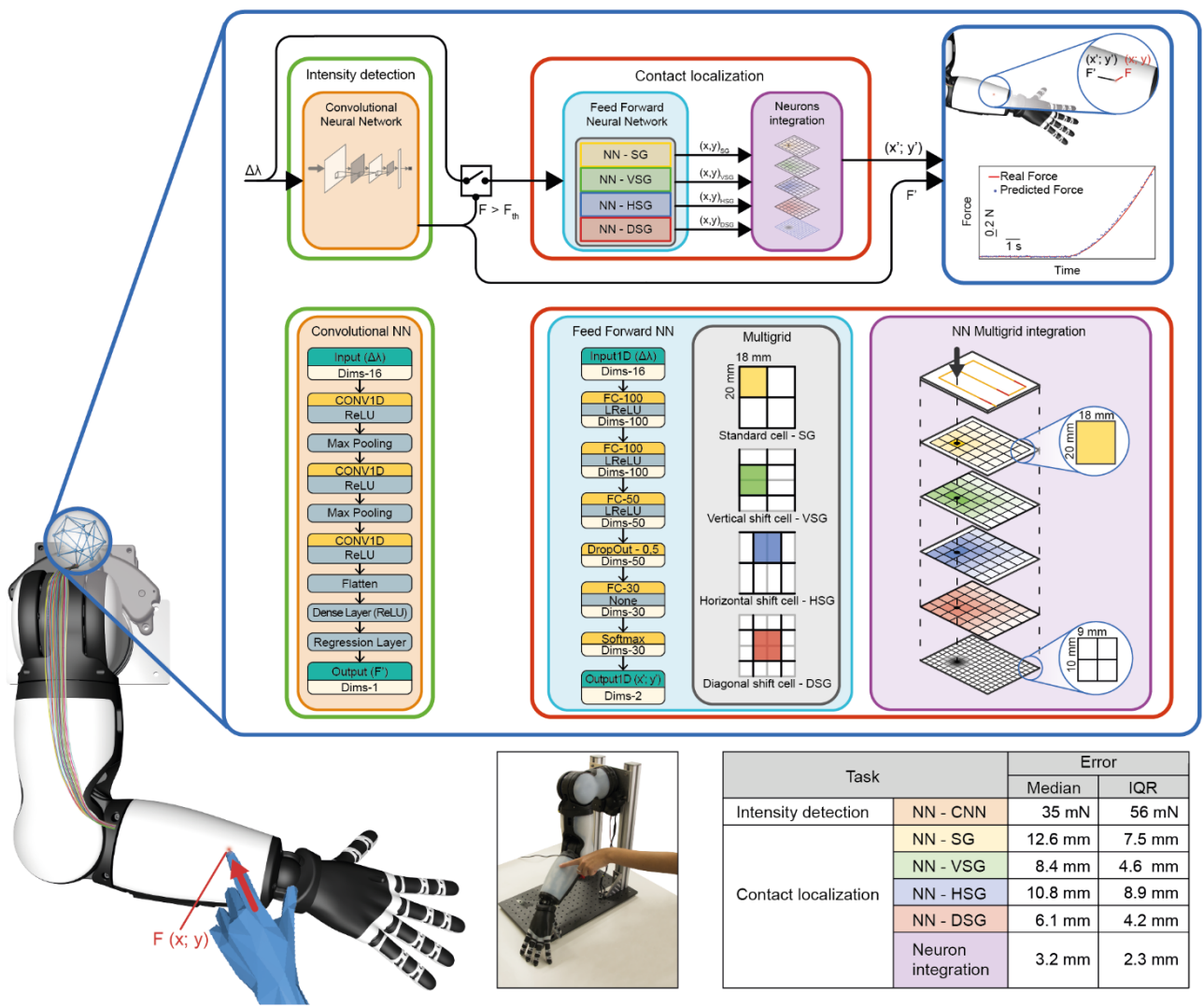

*Fig. 4 | **Deep learning neural network structure and results.** A convolutional neural network is developed for force intensity detection. For forces above the 50 mN threshold, a feed-forward neural network and subsequent multigrid neuron integration process implement the localization of the contact of the load applied onto the skin surface. The table in the bottom right corner shows the main results, namely the error on force intensity detection and on the localization of a stimulus applied onto the skin surface.*

The CNN (flowchart of Fig. 4) was the first step of the processing pipeline (i.e., I*ntensity Detection* block). A 5-folds cross validation was performed and resulted in a median cross-validation error of 30 mN with 1 mN IQR (Extended Data Table 1). Furthermore, a consistent prediction error across the folds was achieved since the median of the interquartile ranges of the single validation fold resulted in 56 mN with 2 mN IQR (Extended Data Table 1). It is worth mentioning that the implemented neural network correctly predicted the force across the whole 0 N – 2.5 N range, as shown in Fig. 4 (upper right) and in the moving average box plot of Fig. 6. In fact, the CNN predictions followed the changes in the force slope occurring during the initial phase of the contact (0.05 N – 0.5 N), as well.

The *Contact Localization* block was the second part of the processing pipeline (Fig. 4), and its activation is contingent upon the *Intensity Detection* block prediction for forces above 50 mN. It consisted of four feed-forward neural networks followed by a multigrid Neuron Integration Process (NIP). More in detail, by relying on the predictions of the individual neural network, each area, that represented the target classes of the neural network, was associated with a confidence level. Therefore, by geometrically overlapping the sub-areas and adding the corresponding weights provided by the individual neural networks, the coordinates of the indentation were predicted. The resulting cross-validation median error was 3.6 mm (IQR = 0.1 mm, see Extended Data Table 2), that is about one fourth of the minimal distance among nearest neighbour sensing elements thanks to triangulation rules learnt by means of DL strategies. The NIP resulted in substantial improvements on the prediction accuracies compared to the single neural networks. Namely, the SG-, DSG-, HSG-, VSG-NN (Extended Data Table 2) resulted in a cross-validation median error of 12.5 mm (IQR = 8.4 mm), 6.5 mm (IQR = 4.4 mm), 11.0 mm (IQR = 9.3 mm) and 8.6 mm (IQR = 4.9 mm).

Extensive training and cross-validation results for both the *Contact Localization* block and the *Intensity Detection* block are presented in Extended Data Table 1 and Extended Data Table 2, respectively.

The test set resulted in a median error of 35 mN (IQR = 56 mN) for force prediction and of 3.2 mm (IQR = 2.3 mm) for position predictions (Fig. 4, bottom right panel). Both the force intensity and the spatial accuracy predictions were robust over the skin surface, although force and localization errors increased at the edges of the skin, because of the relatively limited number of sensors to learn triangulation rules near the boundary (Fig. 5c-d).

The force prediction error increased linearly with the actual force ($R^2$ = 0.97) and by comparing the force absolute error between the force ranges (Fig. 6a-b). Conversely, when the actual force increases, the accuracy in assessing contact position was enhanced in the force range 0.05 N – 0.5 N and stabilized up to 2.5 N (Fig. 6c-d).

In addition, as a benchmark for our ML solution, Random Guess (RG) models, based on available data, were implemented. For the *Intensity Detection RG model*, predictions were set equal to the median of the force in the train set, resulting in a median absolute error of 194 mN (IQR = 714 mN). By comparing it with our CNN solution, via a Wilcoxon signed- rank test, a significant difference was found ($p<0.001$, Cohen's d = 0.84). In the same way, the *Localization* RG model, setting the predictions equal to the position target median from the training set, presented a median absolute error of 34.72 mm (IQR = 24.68 mm). Applying a Wilcoxon signed-rank test between RG model and our model, a significant difference was found ($p<0.001$, Cohen's d = 1.90), thus confirming the effectiveness of the proposed NN-based model for contact localization.

Lastly, we demonstrated real-time capabilities of the Intensity Detection and Contact Localization block via random force-controlled indentations performed on the artificial skin surface. Beyond, we reported how the artificial skin responds appropriately in a real-time framework to the interaction with human impressed forces, providing evidence of potential generalization ability notwithstanding the intrinsic difference in related contact mechanics in comparison to the hemispherical rigid probe used for training the deep learning model for contact localization and force estimation.

## Discussion

The proposed skin fosters HRC using modular tactile patches, that could potentially fit any robot architecture. More specifically, the integrated skin with optical sensors can be used to either cover purposely designed robots or to retrofit existing ones.

Currently, multimodal approaches combining state-of-the-art hardware development and fine sensing skills with advanced AI approaches are showing promising results in artificial skins[60]. Our study targeted this objective, integrating physical and computational intelligence in the presented soft sensitive skin for collaborative robotics, demonstrating the ability to simultaneously predict the location and the intensity of an external load applied on the patch surface. A polymeric matrix embedding FBG transducers was developed and integrated in a human-scale forearm. The functional role of Ruffini corpuscles was a source of bioinspiration in terms of properties of the receptive fields, that physiologically have large single high-sensitivity spots, in contrast to, as an example, the Merkel corpuscles that present multiple smaller hot spots[13]. In this work we showed that most of the FBG sensors integrated in the developed skin exhibited a single large responsive area (FBGs 1, 3-5, 7-11, 14-16, see code published in the Code Ocean repository for details), one featured a single responsive area provided by clustered adjacent subregions (FBG 6) whereas, likely because of local irregularities of the soft encapsulation polymer, a subset exhibited two large responsive areas (FBGs 2, 12 and 13, see published code for details). These results are to some extent comparable to the findings of a background study characterizing the physiological properties of human mechanoreceptors in the forearm, with a single responsive area in 9 out of 10 SA2 mechanoreceptors [13].

Conventional sensing technologies require a wired electrical dipole per each sensor, limiting the number of elements that can be integrated and, thus, the spatial resolution of the large area sensing skins. With respect to these solutions, FBGs offer competitive advantages, such as intrinsic multiplexing capabilities, high sensitivity, ease of dense integration, and immunity to electro-magnetic interference (enabling, e.g., magnetic resonance compatibility in collaborative healthcare applications) [61,62]. Therefore, this approach may be considered a disruptive solution to overcome several constraints within the development of the ideal sensitive skin.

Deep Learning Networks were implemented to retrieve the distributed contact localization (through four MLPs) and the force intensity estimation (through one CNN), starting from the raw FBG wavelengths. Moreover, the NIP strategy permitted to improve the localization results, leveraging the integrated prediction of four MLPs based on half-pitch shifted grids. The median test set error was 35 mN (IQR = 56 mN) and 3.2 mm (IQR = 2.3 mm) for the estimation of the force and the localization of the source, respectively. In particular, the error distribution of localization all over the skin surface (120 mm along Y and 90° along R) and across the whole range of force (0 N – 2.5 N) was uniform, as shown in Fig. 5 and Fig. 6, with lower accuracy and higher error fluctuations observed just at the edges. This suggests to explore the development of a continuous skin in critical applications, rather than discrete patches, so as to remove/mitigate potential boundary effects on sensing accuracy.

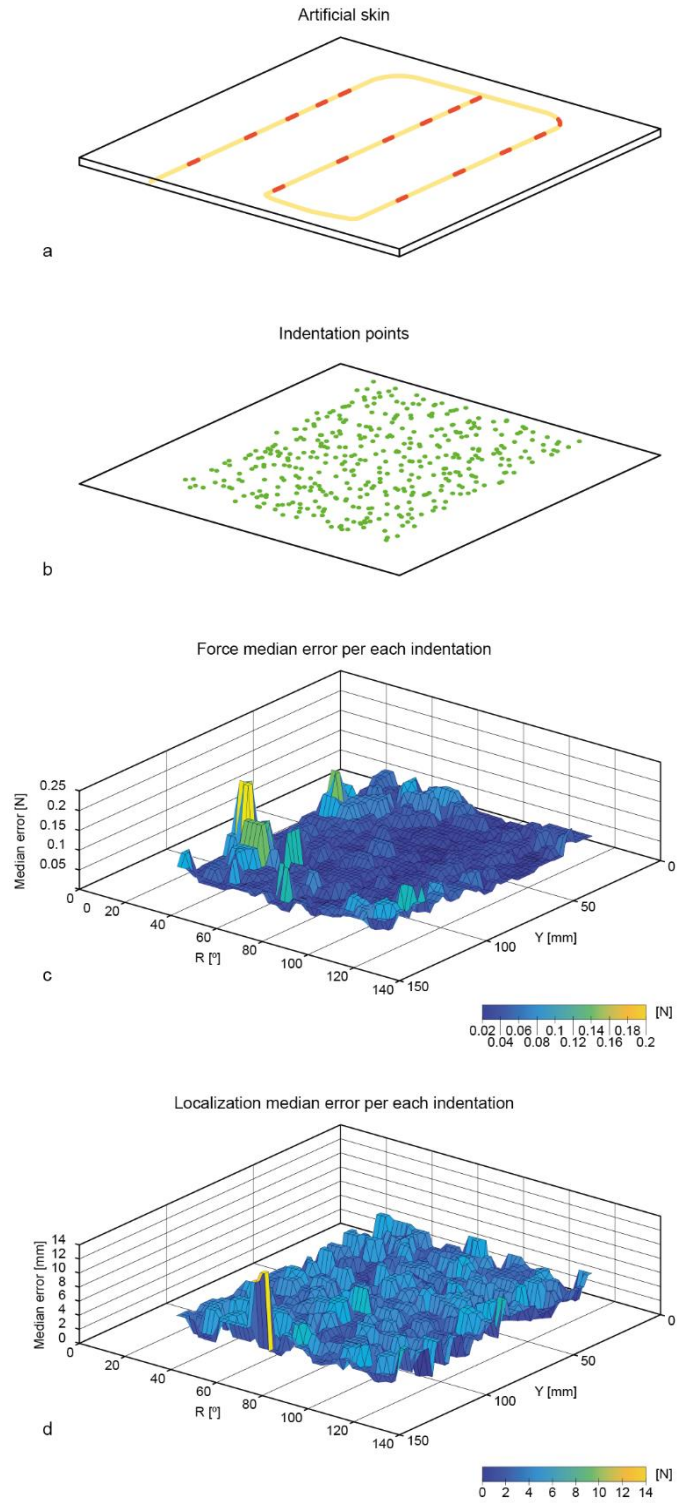

*Fig. 5 | Error distribution across all the indentation test set. a) Path of the optical fiber integrating FBGs into the artificial skin; b) Test indentation points with uniform random distribution over 120 mm in length (Y) and 90° rotation span (R) around the elbow-wrist Y-axis; c-d) Force intensity median error and contact localization median error per each indentation over the whole area, with higher error fluctuations observed just at the edges.*

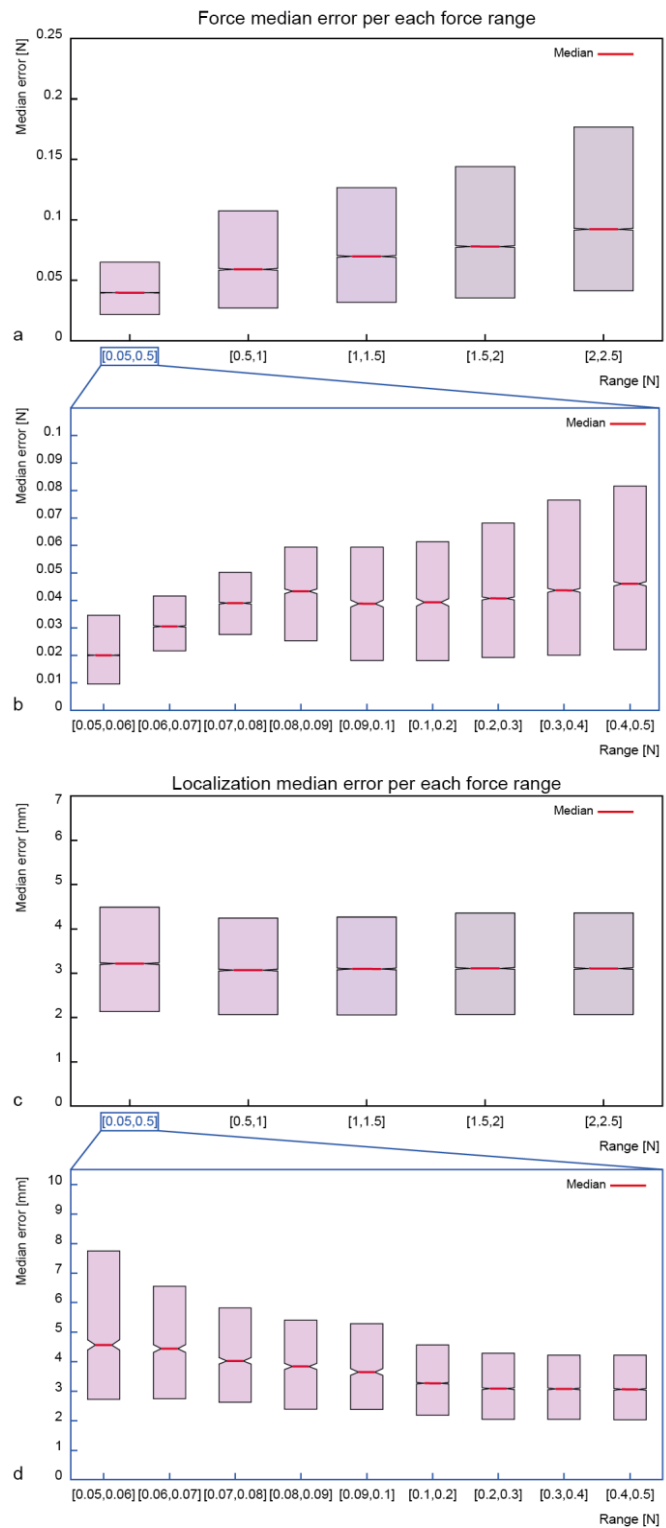

*Fig. 6 | Box plot of the error distribution, showing a monotonic increase of force prediction error as a function of the actual force, and stable position estimation median error.* a) Force median error – range (0.05 N – 2.5 N). b) Inset of the force median error – range (0.05 N – 0.5 N). c) Localization median error – range (0.05 N – 2.5 N). d) Inset of the localization median error – range (0.05 N – 0.5 N).

The present work illustrated multiple breakthroughs with respect to the previous 1D sensing approaches [56]. In particular, a soft tactile skin, made of FBG transducers embedded in a soft matrix, proved that normal load exerted onto its curved and complex surface can be identified in terms of both magnitude and application location. Here, the localization capability throughout the curved and large-area sensorized surface has been demonstrated, whereas in the previous study [56] contact information was retrieved alongside the optical fiber only.

Future works will further investigate generalization ability and robustness of the deep learning strategies, as an example with respect to the possibility to change skin curvature between training and operation phases, and will address the implementation of collaborative behaviours enabled by the availability of such skin patches integrated within a robotic arm to achieve interaction management, path re-planning and robot programming by demonstration.

## Methods

### Biomimetic sensitive e-skin

The skin was a soft-material tactile sensor array system distributed over a large area to enable intensity detection and localization of an external force. Such a skin was designed and developed to cover a 3D-printed, custom-made anthropomorphic robotic arm and to make its forearm sensitive [56,63]. This sensing system consisted of a 8 mm tick stretchable polymeric layer (Dragon Skin 10 Medium, Smooth-on Inc, PA, USA), integrating a 430 mm long optical fiber (FEMTOPlus Grating, FemtoFiberTec GmbH, Berlin, Germany), whose diameter was 80 µm (102 ±5µm with polyimide coating) and bending radius was smaller than 4 mm.

FBGs are micro-resonant structures with a typical length in the range of millimetres, that are inscribed along the core of an optical fiber by means of a laser beam going through a phase mask. The resulting spatial periodic structure is an interference pattern inside the fiber core, which acts as a narrow band optical filter. When a light source illuminates the FBG, part of the light spectrum is transmitted, whereas the residual one, centred around the so-called Bragg wavelength ($\lambda_B$), is reflected backward. The $\lambda_B$ (Fig. 1b), depends on both the effective refractive index of the fiber core ($\eta_{eff}$) and the grating period ($\Delta_B$), or pitch, as defined in Equation (1):

$$\lambda_B = 2\, \eta_{eff}\, \Delta_B \tag{1}$$

FBGs are sensitive to strain, since it affects both $\Delta_B$ and $\eta_{eff}$, so that, the reflected signal ($\lambda_B$) changes accordingly, as shown in Fig. 1b. The efficacy of this technology was previously assessed in robotics, ranging from the simple integration in soft polymers [64–66] to more complex tactile sensors [56] and robotic hand prototypes[67], often in combination with AI techniques and finite element methods [56]. These works share the idea of encapsulating optical fibers in silicone rubbers, for an efficient transfer of the external contact pressures to the embedded FBGs and for mimicking the soft properties of the human skin. The strain experienced by a grating results in a shift of its central wavelength, which is proportional to the distance/intensity from/of the contact interaction. Furthermore, the polymeric substrate enhances the robustness of the encapsulated optical fiber.

The integrated optical fiber was endowed with $n_{FBG}$ = 16 FBGs, each 8 mm in length, $\lambda_B$ ranging from 1530 nm to 1564.5 nm and a pitch of 2.3 nm (pattern shown in Fig. 2d). The skin dimensions were 150 mm along the vertical elbow-wrist axis (Y), covering a 145° region (labelled with R rotational coordinate).

## e-skin fabrication process

The manufacturing process of the soft curved skin for the covering of the anthropomorphic forearm is reported in Extended Data Fig. 3. Several custom moulds were designed and fabricated by means of a 3D printer (Ultimaker S5, Ultimaker, Geldermalsen, Netherlands). A first polymeric layer was casted by pouring Dragon Skin 10 within a mould consisting of the forearm support and a cover with the pattern (1 mm extrusion) for hosting the optical sensor. Once the optical fiber equipped with the 16 FBGs was encapsulated within the dedicated channel, the closing silicone layer was poured into a smooth surface mould.

## FEM simulation

A FEM analysis of the sensitive skin behaviour was performed in COMSOL Multiphysics (COMSOL, Inc., Palo Alto, CA). The simulations addressed the evaluation of a load applied onto the skin top surface when a hemispherical indenter was used. The scope of this preliminary study was to evaluate the effect of an increasing thickness of the encapsulation material, searching for a proper trade-off between the receptive field size and the sensitivity of the sensors. The skin model consisted in a soft polymeric layer with parametric thickness in the range 4mm - 12 mm. The lower limit was set in order to prevent delaminations of the soft polymer and to ease integration of the optical fiber, whereas the upper limit was enforced to guarantee enough space for the actuation units embedded in the robotic arm. In order to simplify the model and reduce the computational burden, a 2D axisymmetric simulation was run, by considering the mid-thickness skin radial strain as representative of that one experienced by the fiber. The polymeric substrate (density = 1070 kg/m$^3$, E = 152 kPa, $\nu$ = 0.49) was modelled as Yeoh hyperelastic material with the following parameter values [68]: $c_1$ = 36 kPa, $c_2$ = 258 Pa, $c_3$ = -0.56 Pa. The steel indenter wad modelled as a linear elastic material, whose mechanical properties were: Young's modulus E = 200 GPa, Poisson coefficient v= 0.30 and density ρ = 7850 kg/m$^3$. The simulation consisted in applying loads by means of the indenter on the top surface of the soft object, replicating the experimental indentations. In the FEM simulation, a null displacement at the bottom surface of the skin was set, since it was attached to the support rigid surface. Moreover, mesh independence was achieved by performing a mesh refinement study until robust results were reached.

## Testing of e-skin sensitivity with Von Frey hairs

The sensitivity of the FBG-based artificial skin was evaluated by means of Von Frey hairs to characterize the human-machine interaction potential. These calibrated microfilaments, widely employed in neurophysiological tests for assessing the human skin sensitivity to mechanical pressure, are nylon filaments with varying diameter: the smaller the diameter, the less force the filament exerts to the skin during their application, before buckling. In particular, 8 Von Frey hairs (60 g, 26 g, 10 g, 4 g, 2 g, 1 g, 0.6 g, and 0.4 g) were selected to manually stimulate the artificial skin. It was asked to a cohort of 12 subjects to provide 20 indentations per each of the 8 filaments (12 subjects x 20 sites x 8 filaments = 1920 stimulations) randomly across the artificial skin surface covering the forearm of the human-like robotic arm (Extended Data Fig. 2a). A graphical user interface (LabView, National Instruments, USA) was developed for data collection and real-time processing, allowing to continuously read and detect the 16 FBG wavelength variations. When at least one sensor output exceeded a wavelength variation threshold, a stimulus was classified as a touch detection. This wavelength threshold was heuristically set to 2 pm for avoiding background noise to determine a spurious contact identification. The stimuli detection rate was then calculated for each microfilament in order to compute a psychometric-like fitting and, thus, the force sensitivity threshold of the artificial skin. Detection rates were fitted by a sigmoid curve:

$$F(x) = \frac{1}{1 + e^{-a(x-b)}}$$

where x is the force associated to the calibrated Von Frey hair, and a and b are the curve steepness and the force value that results in a 50% contact detection probability. The sigmoid coefficients were computed via a non-linear least squares method (Matlab, The Mathworks Inc., USA). The force sensitivity threshold was estimated, as commonly done in psychophysical research [59], by computing the stimulus force value that returns a 75% event probability on the sigmoid fitting.

## Automated testing of the e-skin with a mechatronic platform

A 4 DoF mechatronic platform was used to collect data about position and intensity arising from the force-controlled indentations on the forearm skin. As shown in Extended Data Fig. 4, the apparatus consisted of i) 2 motorized stages (8MTF-102LS05, STANDA, Vilnius, Lithuania; 2.5 µm full step resolution and 102 mm x 102 mm travel range) for horizontal displacements (X-Y), ii) precision motorized positioner (8MVT120-25-4247, STANDA, Vilnius, Lithuania; 5 µm full step resolution and 25.4 mm travel range) for vertical translations (Z), and iii) a motorized rotator (8MR190-2, STANDA, Vilnius, Lithuania; 0.01° full step resolution and 360° rotation range around the Y-axis, resulting in R rotational coordinate). In addition, a 6-axis load cell (Nano-43 with SI-18-0.25 calibration, ATI Industrial Automation, Apex, USA; 1/256 N resolution up to 18 N sensing range), equipped with a cylindrical steel probe (i.e., an indenter of 21 mm in length and with a hemispherical tip of 11 mm in diameter) mimicking the size of a human fingertip, provided the measure of the force component exerted along the loading direction of the skin (Z). A benchmark optical interrogator (FBG-Scan 904, FBGS, Geel, Belgium; 0.3 pm 1σ precision in the 1510 nm – 1590 nm wavelength range) was used to illuminate the FBGs with a broad spectrum and to detect the reflected wavelengths. Data streaming was achieved via the interrogator built-in software (ILLumiSEnse, FBGS, Geel, Belgium). The control of the overall experimental setup and data recording were carried out by means of a dedicated Graphical User Interface (LabVIEW, National Instruments, Texas, USA).

## Automated data collection protocol with mechatronic platform

An experimental protocol consisting of 2700 force-controlled indentations was carried out. Such indentations were spatially randomized throughout the artificial skin surface and performed by means of the steel indenter. Each selected point was loaded up to 2.5 N along the vertical direction (Z) and only data taken during the increasing load phase were retained for further elaborations. The force intensity time series (Fz), the location data (Y; R) and the readouts of the FBG array were collected at a sampling rate of 100 Hz, during each indentation. The total indented area was set to 120 mm along Y and 90° along R and the final dataset included a total of more than 2.7 million samples.

## Evaluation of receptive fields of e-skin FBG sensors

The receptive field of each FBG was assessed by evaluating the spatial distribution of contact force required to achieve a wavelength variation of at least 20 pm. These force levels and each area underpinning them were represented by means of filled 2-D contour plots, where the hot spot regions with maximal sensitivity are represented in black (mapping the spatial regions of the receptive field responsive up to the first sixth of indentation force magnitude).

## Deep learning model and Multigrid Neuron Integration Process

A deep learning model was developed and assessed with the *Ngene LabVIEW* module. The model was made of two main blocks: the *Intensity detection* block, that is a CNN, chosen for its capability of handling time series [69], and the *Localization block*, consisting of 4 dense MLPs. The first block provided the estimation of the force applied onto the skin surface, whereas the second block output estimated contact point coordinates. In particular, when the CNN predicted a force higher than a 50 mN threshold, the *Localization block* was triggered and, after the multigrid *NIP*, the contact point was estimated (Fig. 4). The CNN consisted of 3 identical 3D convolutional layers for features extraction from the raw data. Each of them presented 16 kernel filters (size 16x1x1) and a convolutional stride (S=1), thus matching the input size Rectified Linear Unit (ReLu) activation function [70]. A max-pooling layer followed each convolutional layer and, afterwards, a flattening layer was added. At the end of the stacked convolutional module, two fully-connected layers of 100 and 1 neurons, respectively, were connected to perform regression. The 100 neurons layer responded with a ReLu activation function whilst the last neuron output was implemented via a linear activation function. Then, the $n_{FBG}$ inputs were processed by the second block to perform localization. The 4 classification networks consisted of 3 hidden layers, one dropout layer, for preventing overfitting [71], and an output layer with a softmax activation function.

As regards the Neuron Integration Process, a multigrid strategy was pursued by training four neural networks by using 4 different grid configurations as target classes. Four different virtual grids of 30, 35, 36 and 42 squares (18 mm x 20 mm) were the targets of the classification networks (Fig. 4). With respect to the grid of 30 virtual areas (SG in Fig. 4), the other grids were vertically (VSG), horizontally (HSG) and diagonally (DSG) shifted of half a square, respectively. Each neural network provided a weight, hence a classification percentage, for each area of the corresponding grid.

By virtually subdividing every single square in 4 smaller squares (9 mm x 10 mm) and overlapping the 4 grids, one single finer grid was obtained. The location of the applied load was retrieved with the weighted barycentre of each square.

All indentation tests involving AI strategies were performed on the same sample of sensorized skin, whereas the demonstrations involving the anthropomorphic robotic arm were performed with a different sample.

## Model validation

The implemented validation approach consisted in: 1) spatial random subdivision of the 2700 indentations in training set (85%) and test set (15%), the latter then used to externally evaluate the model performances and 2) a k-fold cross-validation (k = 5) within the training set used to internally validate the model. Models training was implemented via a backpropagation algorithm and Stochastic Gradient Descent optimization algorithm (with momentum 0.9), and the samples were grouped in 50 minibatches to decrease the training time.

It is worth to mention that each indentation was fully assigned, hence not split, to one of the datasets (i.e., training, validation or test set). This choice ensured that no information about the validation/test set was provided during the model training, hence any form of double-dipping was avoided. Moreover, a z-score normalization was applied to both the training set and to the test set, using the mean and standard deviation of the training set.

In addition, to test the algorithm force and position predictions against a random benchmark, Random Guess models were developed. For the *Intensity Detection* RG model, the predictions were set equal to the median force value of the training set (176 mN), whilst for the *Localization* RG model the prediction coordinates were set equal to the median value of the target positions (X = 0 mm and Y = 62.5 mm). Then the absolute error on the test set was computed using the RG model predictions and compared via a two-sided Wilcoxon signed-rank test (on SPSS 27.0, IBM SPSS, Chicago, Illinois, USA) with the CNN and MLP/NIP model test absolute respectively.

## Demonstration of collaborative application

A collaborative application of the developed skin was demonstrated based on its integration into a 7 degrees of freedom (DoFs) robotic arm (3 DoFs for the shoulder, 1 for the elbow, 1 for forearm link pronosupination, and 2 for the wrist). The 2 DoFs of the wrist were actuated with linear motors (P16-50-22-12-P Micro Linear Actuator, Actuonix), which were operated in static position control mode. The remaining 5 DoFs were actuated with servomotors (XM540-W270-R Dynamixel for the shoulder and elbow DoFs, and XH430-W350-R Dynamixel for forearm link pronosupination), which were controlled with two policies. First, the robot arm was in backdrivable configuration and it was freely moved by a human operator, while the joint encoders of the servomotors tracked the operated motion. Afterwards, the recorded trajectories were autonomously replicated by the robot arm, and the skin presented in this study was used in order to detect contact and demonstrate feasibility of collaborative functionality, by temporarily stopping motion in real-time upon contact detection.

## Data availability

Authors are openly sharing test data reported in the manuscript by means of the Code Ocean platform, that can be accessed at the following address: https://codeocean.com/capsule/3018603/tree/v1[72]

## Code availability

Authors are openly sharing the Matlab scripts used to elaborate the dataset and retrieve the figures reported in the manuscript by means of the Code Ocean platform, that can be accessed at the following address:

https://codeocean.com/capsule/3018603/tree/v1[72]

## Acknowledgements

This study was supported in part by the Italian Ministry of Universities and Research through the PARLOMA project (SIN_00132, CMO), by the Italian Ministry of the Economic Development through the Industry 4.0 Competence Center on Advanced Robotics and enabling digital TEchnologies & Systems (ARTES4.0, CMO), by the Tuscany Region through the TUscany NEtwork for BioElectronic Approaches in Medicine: AI based predictive algorithms for fine-tuning of electroceutics treatments in neurological, cardiovascular and endocrinological diseases (TUNE-BEAM, n. H14I20000300002, CMO), and by the European Union's Horizon 2020 research and innovation programme under the Marie Skłodowska-Curie grant agreement No 956745 (European Training Network for InduStry Digital Transformation across Innovation Ecosystems, EINST4INE, 956745, CMO). Results reflect the authors' view only. The funding agencies are not responsible for any use that may be made of the information it contains. CMO gratefully thanks Prof. Silvestro Micera for the comments made to an earlier version of the manuscript.

## Author Contributions Statement

LM, JDA, GT, MZ, EDS and CMO designed and developed the artificial skin embedding the optical fiber. MF, EP, EMS, EDS and CMO contributed to the development of the experimental setup. LM, GF, JDA and MF carried out the experiments and LM, GF, MF and GDA performed data analysis supported by AA, JDA and EP. LM, GF, EDS and CMO conceived and developed the AI algorithms presented in the study. LM and MF performed FEM simulations. CMO conceived the study and was responsible for planning and supervising the scientific work, defining the experimental protocols and of the research grants supporting the study. EDS was co-supervisor of the scientific work, of the definition of the experimental protocols and of FEM simulations. LM, GF, JDA, MF, EDS and CMO wrote the manuscript. AA was responsible for the artworks, the figures and the supplementary materials. All the authors discussed the results, critically revised the paper and approved the final version. Correspondence and requests for materials should be addressed to CMO and EDS.

## Competing Interests Statement

The authors declare the following competing interests: LM, JDA, GT, MZ, EP, EMS, EDS and CMO disclose a patent filed on the developed artificial skin and collaborative robot arm integrating FBG transducers (application number IT201900003657A1). The remaining authors declare no competing interests.

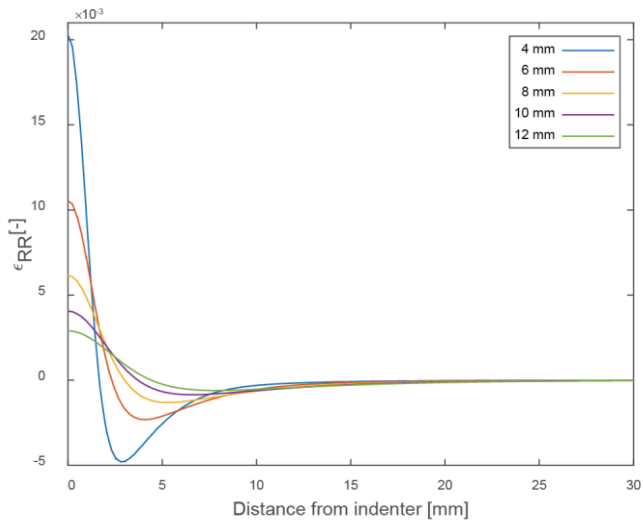

*Extended Data Fig. 1 | Thickness effect of the encapsulation material on the FBG receptive fields and sensitivity within the range 4 to 12 mm at 50 mN.* Higher material thickness reduces sensitivity over the sensor and shifts the curve minimum.

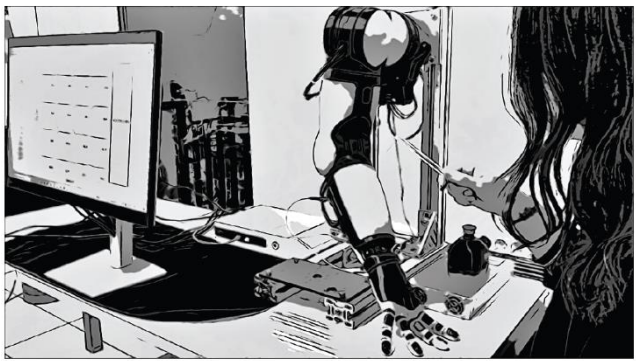
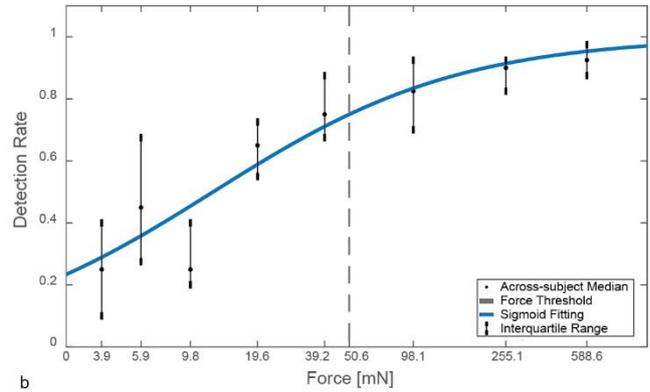

a  b

*Extended Data Fig. 2 | Experiment with 12 subjects involving random administration of calibrated Von Frey hairs over the artificial skin surface to evaluate its detection rate as a function of stimulation force intensity.* a) Illustration of experimental setup and procedure. b) Sigmoid fitting of contact detection rate as a function of the nominal force exerted by the von Frey microfilament. X-axis is represented with logarithmic scale.

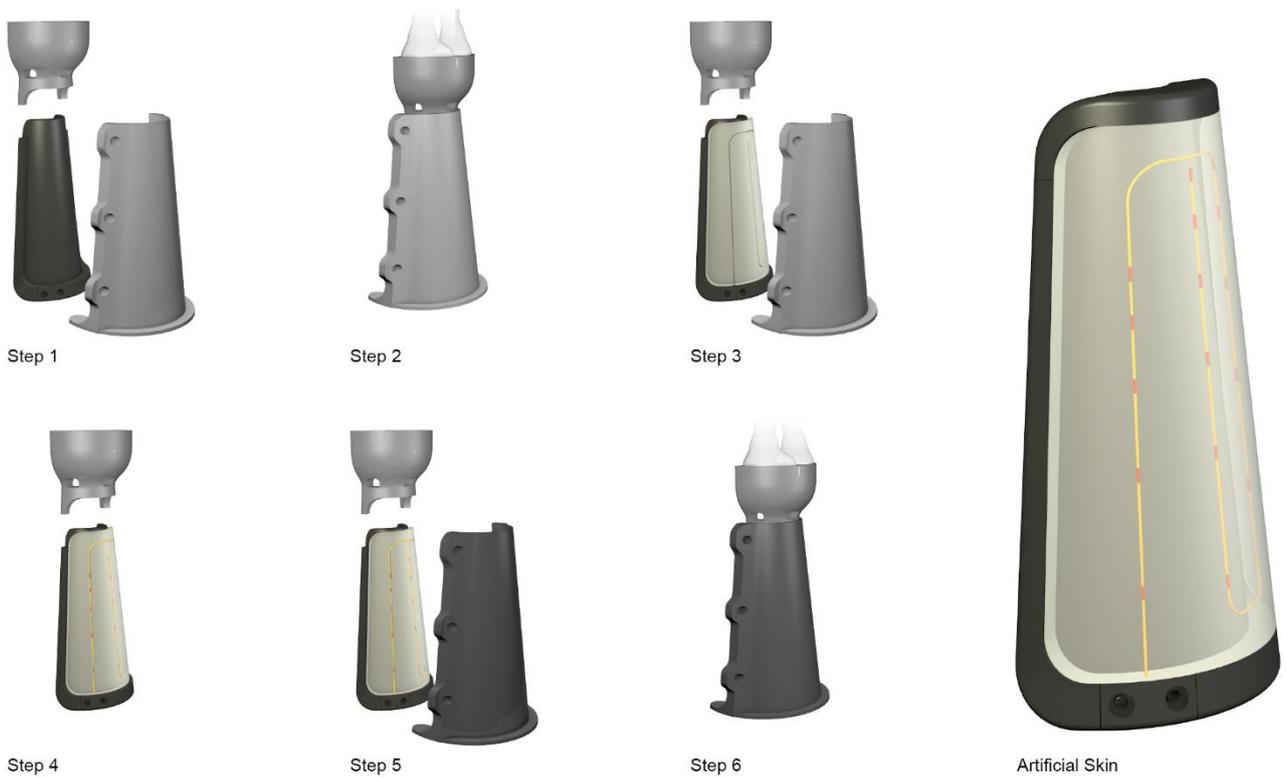

***Extended Data Fig. 3 | Fabrication process of the biomimetic skin.*** *Step 1) Integration of the 1st mold; Step 2) Dragon Skin 10 casting in the 1st mold; Step 3) Removal of the 1st layer of the artificial skin; Step 4) Integration of the optical fiber embedding the FBGs in the soft skin; Step 5) Integration of the 2nd mould; Step 6) Removal of the 2nd layer of the artificial skin and demolding at the end of the procedure.*

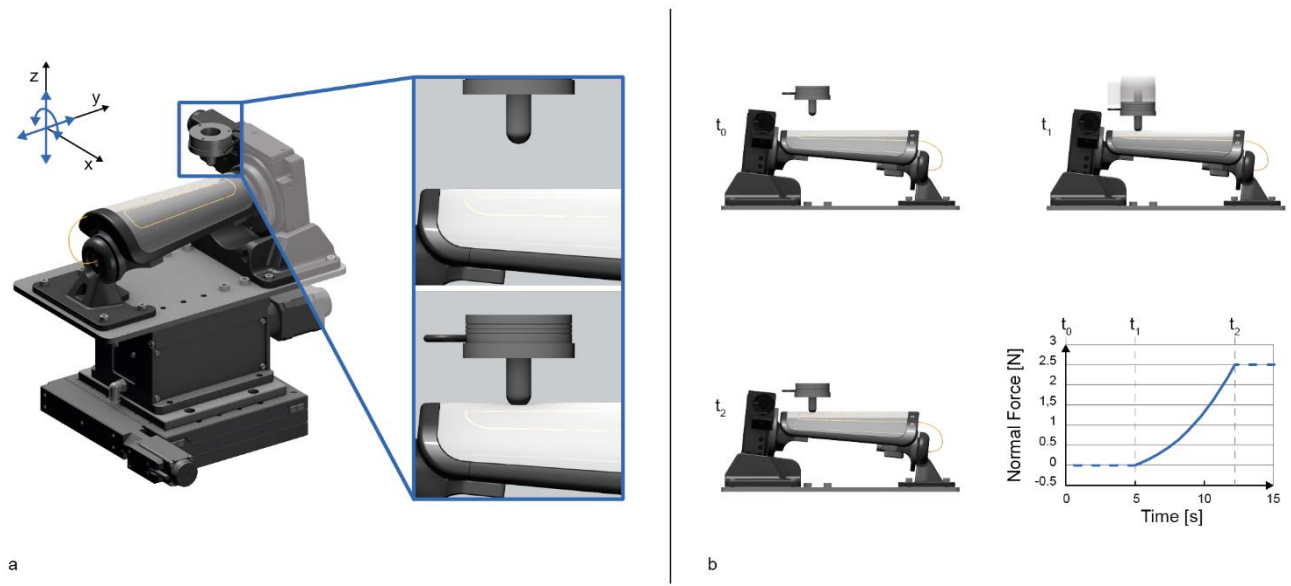

***Extended Data Fig. 4 | Experimental setup and protocol.*** *a) 4-axis mechatronic platform for force-controlled indentations over the skin surface; b) Stimulation force profiles adopted for the indentation experiments (contact force up to 2.5N) to train the neural network.*

*Extended Data Table 1 | CNN 5-fold cross-validation results for force intensity.*

| k-fold | CNN - Force detection Cross-Validation | | | |
|---|---|---|---|---|
| | Training error [mN] | | Validation error [mN] | |
| | Median | IQR | Median | IQR |
| 1 | 33 | 48 | 36 | 53 |
| 2 | 32 | 47 | 35 | 57 |
| 3 | 32 | 48 | 36 | 56 |
| 4 | 31 | 46 | 35 | 57 |
| 5 | 31 | 47 | 35 | 55 |
| Median | 32 | 47 | 35 | 56 |
| IQR | 1 | 1 | 1 | 2 |

*Extended Data Table 2 | MLP 5-fold cross-validation results for contact localization.*

| k-fold | NN | MLP - Localization detection Cross-Validation | | | |
|---|---|---|---|---|---|
| | | Training error [mm] | | Validation error [mm] | |
| | | Median | IQR | Median | IQR |
| 1 | SG-NN | 12.4 | 8.0 | 12.2 | 8.4 |
| | DSG-NN | 6.3 | 4.6 | 6.5 | 4.6 |
| | HSG-NN | 10.9 | 9.3 | 10.7 | 9.6 |
| | VSG-NN | 8.7 | 4.9 | 8.4 | 4.9 |
| | UNION | 3.3 | 2.5 | 3.6 | 2.7 |
| 2 | SG-NN | 12.1 | 8.2 | 12.5 | 7.7 |
| | DSG-NN | 6.2 | 4.6 | 6.7 | 4.9 |
| | HSG-NN | 10.6 | 9.4 | 11.2 | 9.0 |
| | VSG-NN | 8.5 | 4.8 | 8.8 | 4.5 |
| | UNION | 3.4 | 2.5 | 3.6 | 2.6 |
| 3 | SG-NN | 12.5 | 8.0 | 12.3 | 8.5 |
| | DSG-NN | 6.7 | 4.6 | 6.3 | 4.6 |
| | HSG-NN | 10.9 | 9.4 | 11.0 | 9.6 |
| | VSG-NN | 8.6 | 4.7 | 8.6 | 5.2 |
| | UNION | 3.4 | 2.5 | 3.5 | 2.7 |
| 4 | SG-NN | 12.3 | 8.1 | 12.5 | 7.9 |
| | DSG-NN | 6.2 | 4.6 | 6.7 | 4.8 |
| | HSG-NN | 10.9 | 9.5 | 11.0 | 9.2 |
| | VSG-NN | 8.6 | 4.7 | 8.9 | 5.2 |
| | UNION | 3.4 | 2.5 | 3.6 | 2.8 |
| 5 | SG-NN | 12.3 | 8.0 | 12.8 | 8.6 |
| | DSG-NN | 6.3 | 4.6 | 6.3 | 5.0 |
| | HSG-NN | 10.8 | 9.3 | 10.8 | 9.3 |
| | VSG-NN | 8.6 | 4.7 | 8.6 | 4.8 |
| | UNION | 3.4 | 2.4 | 3.4 | 2.5 |
| Median | SG-NN | 12.3 | 8.0 | 12.5 | 8.4 |
| IQR | | 0.2 | 0.1 | 0.3 | 0.7 |
| Median | DSG-NN | 6.3 | 4.6 | 6.5 | 4.8 |
| IQR | | 0.4 | 0.0 | 0.4 | 0.3 |
| Median | HSG-NN | 10.9 | 9.4 | 11.0 | 9.3 |
| IQR | | 0.2 | 0.1 | 0.3 | 0.5 |
| Median | VSG-NN | 8.6 | 4.7 | 8.6 | 4.9 |
| IQR | | 0.1 | 0.1 | 0.3 | 0.5 |
| Median | UNION | 3.4 | 2.5 | 3.6 | 2.7 |
| IQR | | 0.0 | 0.1 | 0.1 | 0.2 |